\documentclass{ieeeaccess}
\usepackage{cite}
\usepackage{amsmath,amssymb,amsfonts}
\usepackage{algorithmic}
\usepackage{graphicx}
\usepackage{textcomp}
\def\BibTeX{{\rm B\kern-.05em{\sc i\kern-.025em b}\kern-.08em
    T\kern-.1667em\lower.7ex\hbox{E}\kern-.125emX}}

\usepackage{soul}
\usepackage{hyperref}
\hypersetup{
    colorlinks=true,
    allcolors=blue
}
\usepackage{floatrow}
\renewcommand{\figureautorefname}{Fig.}
\renewcommand{\tableautorefname}{Table}
\usepackage{booktabs}
\usepackage[caption=false, subrefformat=parens, labelformat=parens]{subfig}
\usepackage{multirow}
\definecolor{ckgreen}{rgb}{0,0.56,0}

\usepackage{array}
\newcommand{\PreserveBackslash}[1]{\let\temp=\\#1\let\\=\temp}
\newcolumntype{C}[1]{>{\PreserveBackslash\centering}p{#1}}
\newcolumntype{R}[1]{>{\PreserveBackslash\raggedleft}p{#1}}
\newcolumntype{L}[1]{>{\PreserveBackslash\raggedright}p{#1}}
\newcommand{\xmark}{$\times$}
\newcommand{\etal}{\textit{et al}.}
\begin{document}
\history{Date of publication xxxx 00, 0000, date of current version xxxx 00, 0000.}
\doi{ZZ.YYYY/ABCDEF.XXXX.WWW}

\title{Less is More: Lighter and Faster Deep Neural Architecture for Tomato Leaf Disease Classification}
\author{\uppercase{Sabbir Ahmed}\authorrefmark{1},
\uppercase{Md. Bakhtiar Hasan}\authorrefmark{1},  \uppercase{Tasnim Ahmed}\authorrefmark{1},
\uppercase{Md. Redwan Karim Sony}\authorrefmark{1},
\uppercase{And Md. Hasanul Kabir}\authorrefmark{1} \IEEEmembership{Member, IEEE}
}
\address[1]{Department of Computer Science and Engineering, Islamic University of Technology, Dhaka, Bangladesh}
% \tfootnote{This paragraph of the first footnote will contain support 
% information, including sponsor and financial support acknowledgment. For 
% example, ``This work was supported in part by the U.S. Department of 
% Commerce under Grant BS123456.''}

\markboth
{Ahmed \headeretal: Lighter and Faster Deep Neural Architecture for Tomato Leaf Disease Classification}
{Ahmed \headeretal: Lighter and Faster Deep Neural Architecture for Tomato Leaf Disease Classification}

\corresp{Corresponding author: Md. Bakhtiar Hasan (e-mail: \hyperlink{mailto:bakhtiarhasan@iut-dhaka.edu}{bakhtiarhasan@iut-dhaka.edu}).}

\begin{abstract}
To ensure global food security and the overall profit of stakeholders, the importance of correctly detecting and classifying plant diseases is paramount. In this connection, the emergence of deep learning-based image classification has introduced a substantial number of solutions. However, the applicability of these solutions in low-end devices requires fast, accurate, and computationally inexpensive systems. This work proposes a lightweight transfer learning-based approach for detecting diseases from tomato leaves. It utilizes an effective preprocessing method to enhance the leaf images with illumination correction for improved classification. Our system extracts features using a combined model consisting of a pretrained MobileNetV2 architecture and a classifier network for effective prediction. Traditional augmentation approaches are replaced by runtime augmentation to avoid data leakage and address the class imbalance issue. Evaluation on tomato leaf images from the PlantVillage dataset shows that the proposed architecture achieves $99.30\%$ accuracy with a model size of $9.60$MB and $4.87$M floating-point operations, making it a suitable choice for low-end devices. Our codes and models are available at \href{https://github.com/redwankarimsony/project-tomato}{https://github.com/redwankarimsony/project-tomato}.
\end{abstract}


\titlepgskip=-15pt

\maketitle

\section{Introduction}
\label{sec:introduction}
\PARstart{T}{omato}, \textit{Solanum lycopersicum}, is one of the most common vegetables grown worldwide. According to recent statistics, around 180.64 million metric tons of tomatoes are grown worldwide which amounts to an export value of 8.81 billion US Dollars \cite{tridge2021tomato}. However, the production of tomatoes is on the decline due to the crop being prone to various diseases \cite{panno2021review}. Traditional disease detection approaches require manual inspection of diseased leaves through visual cues or chemical analysis of infected areas, which can be susceptible to low detection efficiency and poor reliability due to human error. Adding to the problem, the lack of professional knowledge of the farmers and the unavailability of agricultural experts who can detect the diseases in remote areas also hamper the overall harvest production. Negligence in this regard poses a significant threat to food security worldwide while causing great losses for the stakeholders involved in tomato production. Early detection and classification of diseases implemented using tools and technologies available to the farmers can go a long way to alleviate all the issues discussed \cite{li2021plant}.

Several solutions have been proposed using the traditional machine learning approaches for plant disease classification \cite{liakos2018machine}. Moreover, the emergence of deep learning-based methods in the agricultural domain has opened a new door for researchers with outstanding generalization capability removing the dependencies on handcrafted features \cite{kamilaris2018deep}. Recently, Convolutional Neural Network (CNN) has become a powerful tool for any classification task as it automatically extracts important features from images without human supervision. Moreover, the recent variations of CNN architectures such as 
AlexNet \cite{krizhevsky2012imagenet}, 
DenseNets \cite{huang2017densely},
EfficientNets \cite{tan2019efficientnet},
GoogLeNet \cite{szegedy2015going}, 
MobileNets \cite{howard2017mobilenets,sandler2018mobilenetv2}, 
NASNets \cite{zoph2018learning},
Residual Networks (ResNets) \cite{he2016deep},
SqueezeNet \cite{iandola2016squeezenet},
Visual Geometric Group (VGG) Networks \cite{simonyan2015very} have enabled the machines to understand complex patterns enabling even better performance than humans in many classification problems. 

With the introduction of transfer learning where the reuse of a model efficient in solving one problem as the starting point of another problem in a relevant domain has significantly reduced the requirement of vast computational resources \cite{torrey2010transfer}. Consequently, the utilization of pretrained AlexNet and GoogLeNet architectures on the publicly available PlantVillage Dataset \cite{hughes2015open} has been one of the pioneer works of leaf disease classification using transfer learning and paved the way for numerous solutions in the existing literature \cite{mohanty2016using}. These deep neural architectures have been found to be extremely helpful for leaf disease classification tasks for several plants such as, apple \cite{zia2021recognizing}, cassava \cite{abayomiAlli2021CassavaDR}, corn \cite{waheed2020anOptimized}, cucumber \cite{zhang2019cucumber}, grape \cite{liu2020aData}, maize \cite{zhang2018identification}, mango \cite{singh2019multilayer}, rice \cite{lu2017indentification}, etc. However, most of these solutions propose deep and complex networks focusing on increasing the accuracy of detection.

Real-life applications, such as agriculture, often require small and low latency models tailored explicitly for devices with small memory and low computational power while also having comparable, if not better, accuracy. Most of the systems focusing on lightweight models had to sacrifice accuracy and/or work with a limited number of diseases/samples. In this regard, this work proposes a lightweight and fast deep neural architecture for tomato leaf disease classification. The system utilizes a pretrained MobileNetV2 as a feature extractor followed by a classifier network. Contrast Limited Adaptive Histogram Equalization (CLAHE) technique has been used to reduce the effect of poor lighting conditions from the leaf images and enhance the disease spots without increasing the noise.
We have tackled the dataset imbalance, overfitting, and data leaking issues by applying runtime augmentation in different dataset splits. 

The performance of the model has been evaluated on tomato leaf images from the PlantVillage dataset incorporating a healthy and nine disease classes. The baseline accuracy of the MobileNetV2 architecture was 97.27\%. The proposed pipeline was robust enough to uplift it to 99.30\%. Further comparison with the state-of-the-art tomato leaf disease classification models showed that the proposed approach is competent enough to achieve high accuracy while maintaining a relatively small model size and reduced number of computations. The nearest model producing a similar level of performance required a 2.4 times heavier model with 2.45 times additional FLOPs count requirements. Hence, this approach can pave the way for a suitable solution for designing real-life applications in low-end devices available to the farmers.

The rest of the paper is organized as follows. We discuss the research gaps in the existing literature to justify our motivation in Section {\ref{sec:related}}. The details of our proposed methodology are described in Section {\ref{sec:materials}}. Our experimental results justifying the efficacy of our proposed method is provided in Section {\ref{sec:result}}. And finally, in Section {\ref{sec:conclusions}}, we deliver our concluding remarks, discuss our limitations, and provide directions for future researchers.

\section{Related Works}
\label{sec:related}
Current research trends on tomato leaf disease classification tend to focus on developing solutions using Deep Neural Architectures, simplifying networks for faster computation targeting embedded systems, real-time disease detection, etc. The introduction of intelligent systems incorporating these solutions could go a long way to reduce crop yield loss, remove tedious manual monitoring tasks, and minimize human efforts.

Earlier approaches in tomato leaf disease classification involved different image-based hand-crafted feature extraction techniques that were fed into machine learning-based classifiers. These works mainly focused on only a few diseases with extreme feature engineering and were often limited to constrained environments. 
%%% About Feature Extractors:
To extract features, researchers focused on utilizing different image-level feature extraction techniques like Gray-Level Co-occurrence Matrices (GLCM) \cite{mokhtar2015svm}, Geometric and histogram-based features 
\cite{mokhtar2015identifying}, Gabor Wavelet Transformation \cite{mokhtar2015tomato}, Moth-Flame Optimization and Rough Set (MFORS) \cite{hassanien2017improved}, and similar techniques.
%Some works extracted the Region of Interest (RoI)
To segment the diseased portion of the leaves, several works have extracted the Region of Interest (RoI) using k-means clustering \cite{mokhtar2015identifying}, Otsu's method \cite{sabrol2016tomato}, etc.
%%% About Classifiers: 
To predict the class labels from the extracted features, Support Vector Machine (SVM) \cite{mokhtar2015svm, mokhtar2015tomato}, Decision Trees \cite{sabrol2016tomato}, and other classifiers were used.
%%% Why ML approaches cannot sustain:
Due to their sensitivity to the surroundings of leaf images, machine learning approaches relied on rigorous preprocessing steps like manual cropping of RoI, color space transformation, resizing, background removal, and image filtering for successful feature extraction. This increased complexity due to preprocessing limited the traditional machine learning approaches to classify a handful of diseases from a small dataset, thus failing to generalize on larger ones.

%%% ML to MISC to DL connection
The performances of a significant portion of the prior works were not comparable as they were mostly done on self-curated small datasets. This issue was alleviated to a great extent when the PlantVillage dataset was introduced containing 54,309 images of 14 different crop species and 26 diseases \cite{hughes2015open}.
%One of the pioneering works in this dataset by \cite{mohanty2016using} demonstrated how deep CNN models can outperform the traditional shallow models in plant leaf disease detection.  
A subset of this dataset contains nine tomato leaf diseases and one healthy class that has been utilized by most of the recent deep learning-based works on tomato leaf disease classification.
Several works on tomato leaf diseases also focused on segmenting leaves from complex backgrounds \cite{ngugi2020tomato}, real-time localization of diseases \cite{liu2020tomato, zhang2020deep, fuentes2017robust}, detection of leaf disease in early-stage \cite{liu2020early}, visualizing the learned features of different layers of CNN model \cite{brahimi2017deep, fuentes2017spectral}, combining leaf segmentation and classification \cite{yang2020self}, and so on. These works mostly targeted removing the restrictions of lighting conditions and uniformity of complex backgrounds.

\begin{figure*}[tb]
    \centering
    \includegraphics[width=\textwidth]{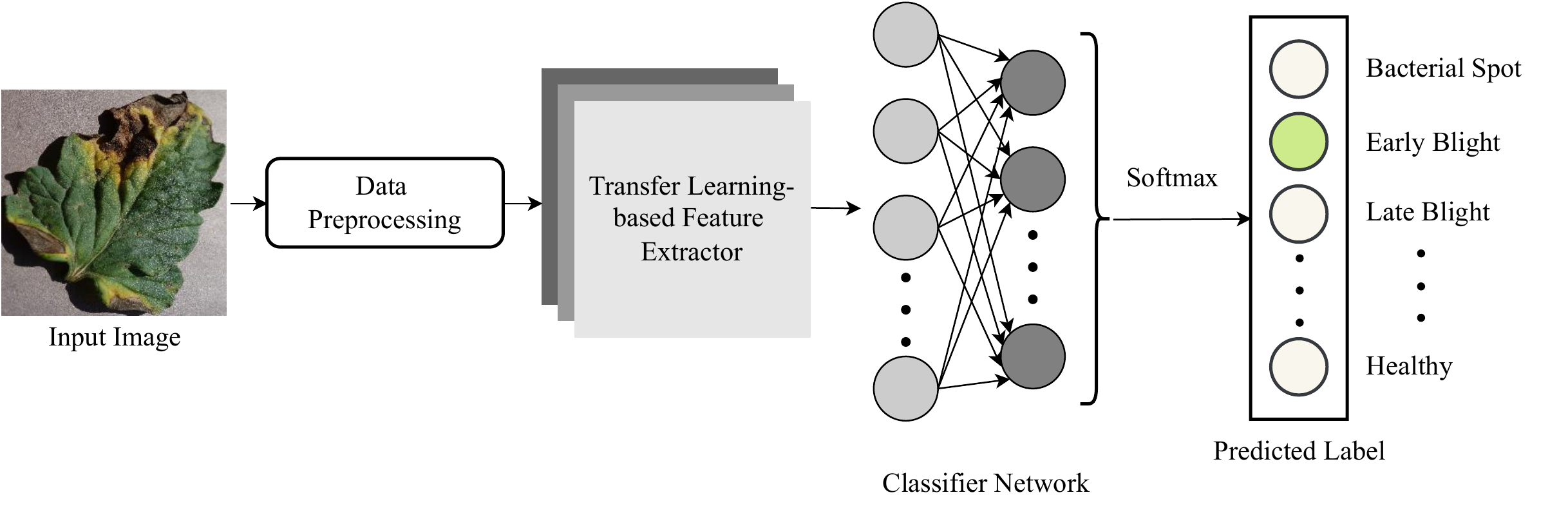}
    \caption{Overview of the tomato leaf disease classification architecture}
    \label{fig:overview}
\end{figure*}

% Entry of TL methods0
To alleviate the dependency on hand-crafted features along with achieving better classification accuracy with large datasets, recent transfer learning-based approaches to leaf disease classification have investigated the performance of different pretrained models using various hyperparameters.
Based on their results, they recommended the use of GoogleNet \cite{brahimi2017deep, maeda2020comparison, wu2020dcgan}, AlexNet \cite{rangarajan2018tomato}, ResNet \cite{zhang2018can}, DenseNet121 \cite{abbas2021tomato} in creating tomato leaf disease detection systems due to their superior performance compared to other models. Some of these works have also investigated the effect of different hyperparameter choices like optimizers, batch sizes, the number of epochs, and fine-tuning the model from different depths to see how they impact its performance \cite{maeda2020comparison, rangarajan2018tomato}.
These models were pretrained on massive datasets, making them the perfect choice for extracting relevant features outperforming shallow machine learning-based models. 
Although these systems achieved high accuracy going up to $99.39\%$ \cite{maeda2020comparison}, the models were huge and computationally expensive, often making them infeasible for low-end devices. 

%with limited memory and processing power.

% \begin{center}
% \bxfigure[h]{\label{fig:overview}Overview of the tomato leaf disease classification architecture}{
%  \includegraphics[width=0.9\textwidth]{ahmed1.pdf}
% }
% \end{center}

%%% Solutions for low-end devices
Several attempts were made to reduce the computational cost and model size. Durmuc\c{s} \etal \cite{durmucs2017disease} utilized SqueezeNet to detect tomato leaf diseases. The base SqueezeNet architecture reduces the computational cost by minimizing the number of $3\times3$ filters, late downsampling, and deep compression. The authors conducted the experiments on an NVIDIA Jetson TX1 device targeting real-time disease detection using robots.
Tm \etal \cite{tm2018tomato} proposed a variation LeNet, one of the earliest and smallest deep-learning architectures. The authors introduced an additional convolutional and pooling layer to the base architecture and increased the number of filters in different layers to extract complex features. However, the accuracies achieved by these two systems were not on par with the performance of the deeper models.
Bir \etal \cite{bir2020transfer} utilized pretrained EfficientNet-B0 to achieve a comparable accuracy with the state-of-the-art while keeping the model size and computation low. This architecture applies grid search to find coefficients for width, depth, and resolution scaling to reduce the size of the baseline model with a minimal impact on accuracy. However, when classifying the tomato leaves, the authors had to discard a significant number of samples to gain a comparable accuracy. Reduction of dataset size in this manner, even if balanced with augmentation, might result in discarding complex samples restricting the generalization capability of the models. 
%%This leads to the necessity of lightweight models that can achieve state-of-the-art performance with high generalization capability.
All these issues impose the requirement of lightweight models that can achieve state-of-the-art performance with high generalization capability.

\section{Materials and Methods}
\label{sec:materials}
Our proposed architecture takes tomato leaf images as input and outputs the class labels. At first, the input image is passed through a preprocessing step where it is enhanced using Adaptive Histogram Equalization.
Then, the enhanced image is fed to a transfer learning block, where we utilize a pretrained deep CNN model for efficient feature extraction. 
To determine a suitable feature extractor, we experimented with nine different pretrained architectures which are DenseNet121, DenseNet201, EfficientNet-B0, MobileNet, MobileNetV2, NASNet-Mobile, ResNet50, ResNet152V2, and VGG19. Based on the results, we have chosen MobileNetV2 due to its smaller size and faster inference while maintaining comparable accuracy.
Then the features extracted by the pretrained model are fed through a shallow densely connected classifier network to get the Softmax probabilities for every class using which we predict the final label. The general pipeline of the proposed approach is depicted in \figureautorefname~\ref{fig:overview}.

\subsection{Dataset}
\begin{figure*}[htb]
    \centering
    \subfloat[Bacterial Spot]{
    \includegraphics[width=0.17\textwidth]{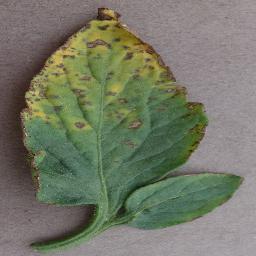}	
    }
    \subfloat[Early Blight]{
    \includegraphics[width=0.17\textwidth]{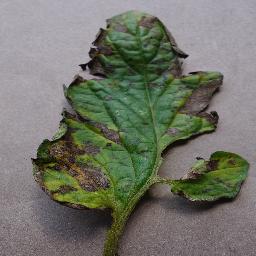}	
    }
    \subfloat[Late Blight]{
    \includegraphics[width=0.17\textwidth]{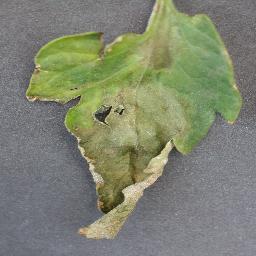}	
    }
    \subfloat[Leaf Mold]{
    \includegraphics[width=0.17\textwidth]{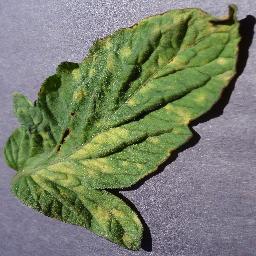}	
    }
    \subfloat[Septorial Leaf Spot]{
    \includegraphics[width=0.17\textwidth]{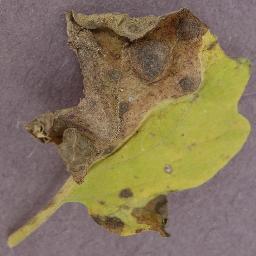}	
    }\\
    \subfloat[Two-spotted Spider Mites]{
    \includegraphics[width=0.17\textwidth]{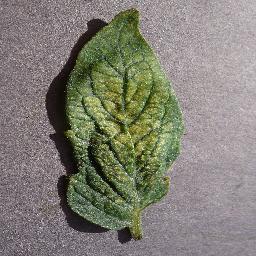}	
    }
    \subfloat[Target Spot]{
    \includegraphics[width=0.17\textwidth]{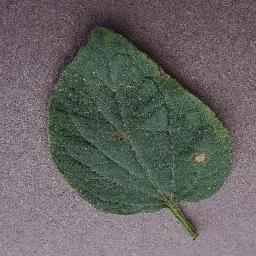}	
    }
    \subfloat[Yellow Leaf Curl Virus]{
    \includegraphics[width=0.17\textwidth]{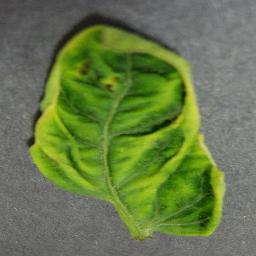}
    }
    \subfloat[Tomato Mosaic Virus]{
    \includegraphics[width=0.17\textwidth]{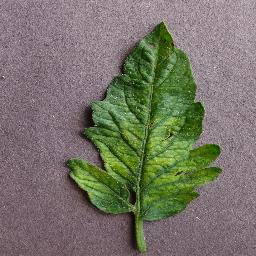}	
    }
    \subfloat[Healthy]{
    \includegraphics[width=0.17\textwidth]{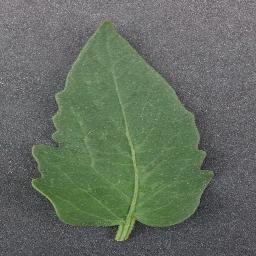}	
    }
    
    \caption{Sample tomato leaf images of the 10 classes from the PlantVillage dataset}
    \label{fig:tomdis}
\end{figure*}

As of today, the PlantVillage Dataset\footnote{\href{https://www.tensorflow.org/datasets/catalog/plant_village}{https://www.tensorflow.org/datasets/catalog/plant\_village}} is the largest open-access repository of expertly curated leaf images for disease diagnosis. The dataset comprises 54,309 images of healthy and infected leaves belonging to 14 crops, labeled by plant pathology experts.
Among them, 18,160 images are of tomato leaves, divided into one healthy and nine disease classes. 
This dataset offers a wide variety of diseases and contains samples of leaves being infected by various diseases to different extents.
One sample image from each class can be seen in \figureautorefname~\ref{fig:tomdis}.

\begin{table}[bt]
    \centering
    \caption{Distribution of samples in the dataset}
    \label{tab:balance}
    
    \begin{tabular}{L{0.5\columnwidth} C{0.3\columnwidth}}
        \toprule
        Class Label & Sample Count\\\midrule
        Bacterial Spot & $2127$\\
        Early Blight  & $1000$\\
        Late Blight  & $1909$\\
        Leaf Mold & $952$\\
        Septoria Leaf Spot & $1771$\\
        Two-spotted Spider Mites & $1676$\\
        Target Spot & $1404$\\
        Yellow Leaf Curl Virus & $5357$\\
        Tomato Mosaic Virus & $373$\\
        Healthy  & $1591$\\
        \midrule
        Total & $18160$\\
        \bottomrule
    \end{tabular}
\end{table}

From the distribution of the number of samples in different classes shown in \tableautorefname~\ref{tab:balance}, it is evident that the dataset contains an imbalance as different classes have a significantly varying number of samples. The maximum number of samples is $5357$, belonging to Yellow Leaf Curl Virus disease, whereas the number of samples corresponding to Mosaic Virus disease is as low as $373$. Few problems arise because of this class imbalance. First, the model does not get a good look at the images of classes with a lower number of samples, leading to less generalization \cite{chawla2002smote}. Moreover, the overall accuracy might still be high even if the model is ignoring these small-sized classes, as they do not contribute much to the overall accuracy  \cite{leevy2018survey}. Different techniques involving undersampling and oversampling can be employed to tackle this issue, ensuring that the model is equally capable of identifying all diseases.

\subsection{Data Preprocessing}
Disease spots often have close intensity values with the surroundings due to the poor lighting condition of the images provided in the dataset.
Moreover, in real-world applications, images captured by the end-users might not always be adequately illuminated, and this might fail to provide the model with enough details to identify the disease, and hence affect the classification result \cite{li2016underwater}. Contrast enhancement techniques like histogram equalization can be applied to enhance the details and correct the illumination problem. Generally, histogram-based approaches work globally throughout the image. However, the intensity distribution of the leaf regions can be different from that of the background. So, the same transformation function cannot be applied to the entire image. To tackle the illumination problem by addressing the uneven distribution of intensity, we opted for Contrast Limited Adaptive Histogram Equalization \cite{pizer1987adaptive}.

Furthermore, as mentioned earlier, there exists a class imbalance in the original dataset. This issue has been tackled in various ways in the existing literature. The most common way of dealing with this has been to undersample and/or oversample certain classes \cite{zhang2018can, bir2020transfer, wu2020dcgan, abbas2021tomato}. Although it makes the dataset balanced to some extent, it has its own drawbacks. Undersampling may drop some of the challenging images for certain classes that can contain important information for the model to learn, which eventually hinders the generalizing capability of the model. Oversampling utilizes different data augmentation techniques to produce multiple copies of the original images, each having slight variations. But if we perform augmentation before splitting the dataset into train, validation, and test sets, it might inject slight variations of the training set into the test set. As the model learns to classify one variation of the image while training, it is highly likely to correctly classify the other variations in the test set, overestimating the accuracy of the system. This problem is known as data leakage \cite{kaufman2012leakage}. As each choice has its pros and cons, we decided to perform data augmentation during runtime.

\subsubsection{Contrast Limited Adaptive Histogram Equalization (CLAHE)}

\begin{figure}[htb]
	\centering
	\subfloat[Original Image\label{fig:clahe1}]{
	\includegraphics[width=.4\columnwidth]{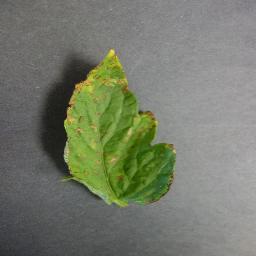}
	}
	\subfloat[Enhanced Image\label{fig:clahe2}]{
	\includegraphics[width=.4\columnwidth]{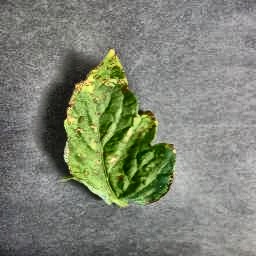}	
	}
	\caption{Illumination correction using Contrast Limited Adaptive Histogram Equalization.}
	\label{fig:CLAHE}
\end{figure}

CLAHE increases the contrast between diseased spots and the leaf by dividing the image into multiple small regions and applying a transformation function that is proportional to a cumulative distribution function. This function is calculated based on the histogram of the intensity distribution of the pixels inside each region. CLAHE also limits the amplification of the noise, which is prevalent in low light images, near regions with constant intensity by clipping the histogram value beyond a threshold. \figureautorefname~\ref{fig:CLAHE} shows the sample output after applying CLAHE on an original image.

Before applying CLAHE, the leaf image was converted from RGB color space to Hunter Lab color space. The intensity channel of the leaf image was divided into $P\times Q$ regions, where $P$ denotes the number of contextual regions on the x-axis, and $Q$ denotes the number of contextual regions on the y-axis. Our empirical results demonstrated that a value of $7$ for both $P$ and $Q$ provided the best results.

A histogram was calculated considering the intensity of the pixels for each contextual region. Then each histogram was clipped based on a threshold $\beta$, which was set to be $3$ upon experimentation. After that, using the Cumulative Distribution Function, a function was generated to map the input intensities with the desired output intensities. While mapping, the function performed Bilinear Interpolation of the four nearby regions to reduce the blocking effect. Finally, the leaf image was converted from Hunter Lab color space to RGB color space.

To maintain consistency, we have preprocessed all the tomato leaf images of the dataset using CLAHE before feeding them to the model. 

\subsubsection{Data Augmentation}
\label{subsubsection:dataAug}

To reflect real-life scenarios, we have picked height and width shifting, clockwise and counterclockwise rotation, shearing, and horizontal flipping to augment the leaf images during runtime.

\begin{figure}[htb]
	\centering
	\subfloat[Enhanced Image\label{fig:enhanced}]{
	\includegraphics[width=0.3\columnwidth]{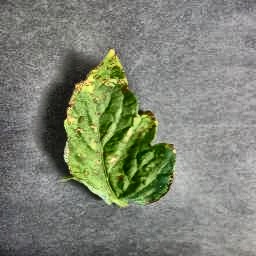}	
	}
	\subfloat[Height Shift\label{fig:height}]{
	\includegraphics[width=0.3\columnwidth]{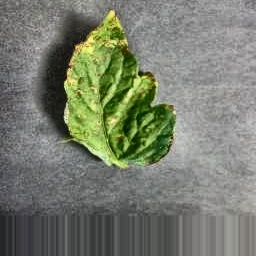}	
	}
	\subfloat[Width Shift\label{fig:width}]{
	\includegraphics[width=0.3\columnwidth]{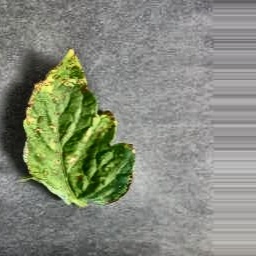}	
	}\\
	\subfloat[Rotation\label{fig:rotate}]{
	\includegraphics[width=0.3\columnwidth]{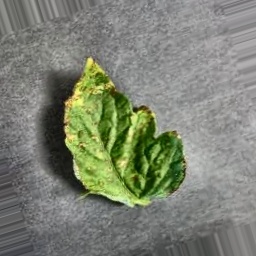}	
	}
	\subfloat[Shearing\label{fig:shearing}]{
	\includegraphics[width=0.3\columnwidth]{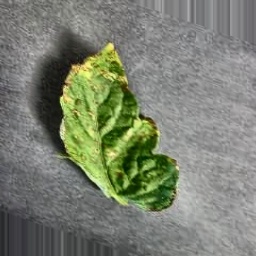}	
	}
	\subfloat[Horizontal Flip\label{fig:horizontal}]{
	\includegraphics[width=0.3\columnwidth]{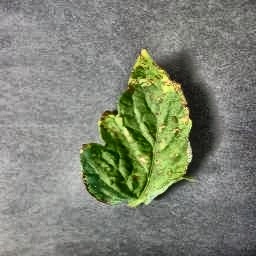}	
	}
	\caption{Data augmentations. A combination of these augmentations were applied randomly during run-time.}
	\label{fig:aug}
\end{figure}

Height and Width Shifting is performed by translating each pixel of the image respectively in the horizontal and vertical direction by a constant factor. In our case, the constant factor was chosen randomly within the range $[0, 0.2]$. While shifting, the pixels going outside the boundary are discarded, and the empty regions are filled with the RGB values of the nearest pixels. \figureautorefname~\ref{fig:height} and \ref{fig:width} shows the effect of performing height and width shift, respectively.

Rotation is performed with respect to the center pixel of the image. For our case, the rotation angle was chosen randomly within the range $[\text{-}20, 20]$ degrees. \figureautorefname~\ref{fig:rotate} shows the effect of performing rotation. 
Shearing is performed by moving each pixel towards a fixed direction by an amount proportional to the pixel's distance from the bottom-most pixels of the image based on a shearing factor. We randomly picked the shearing factor within the range $[0, 0.2]$. \figureautorefname~\ref{fig:shearing} shows the effect of performing shearing.
Flipping an image horizontally requires mirroring the pixels with respect to the centerline parallel to the x-axis. \figureautorefname~\ref{fig:horizontal} shows the effect of performing horizontal flipping.

\begin{figure}[htb]
    \centering
    \subfloat[\label{fig:randAug1}]{
	\includegraphics[width=0.3\columnwidth]{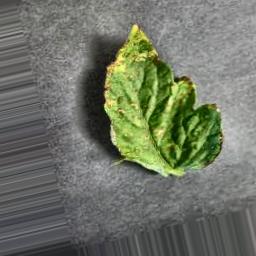}
	}
	\subfloat[\label{fig:randAug2}]{
	\includegraphics[width=0.3\columnwidth]{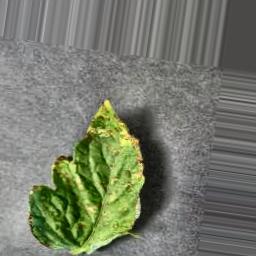}
	}
	\subfloat[\label{fig:randAug3}]{
	\includegraphics[width=0.3\columnwidth]{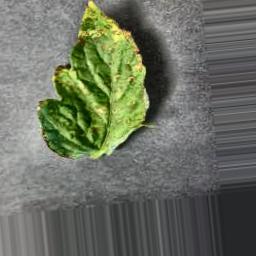}
	}
    \caption{Sample augmentations performed on the images during training, validation, and testing phase}
    \label{fig:randaug}
\end{figure}

Multiple random augmentations are applied to the same image to ensure that the model sees a new variation on every epoch and thus learns to recognize a variety of images. \figureautorefname~\ref{fig:randaug} shows the effect of combining different augmentations that are used during the training, validation, and testing phase.

\begin{figure*}[tb]
    \centering
    \includegraphics[width=\textwidth]{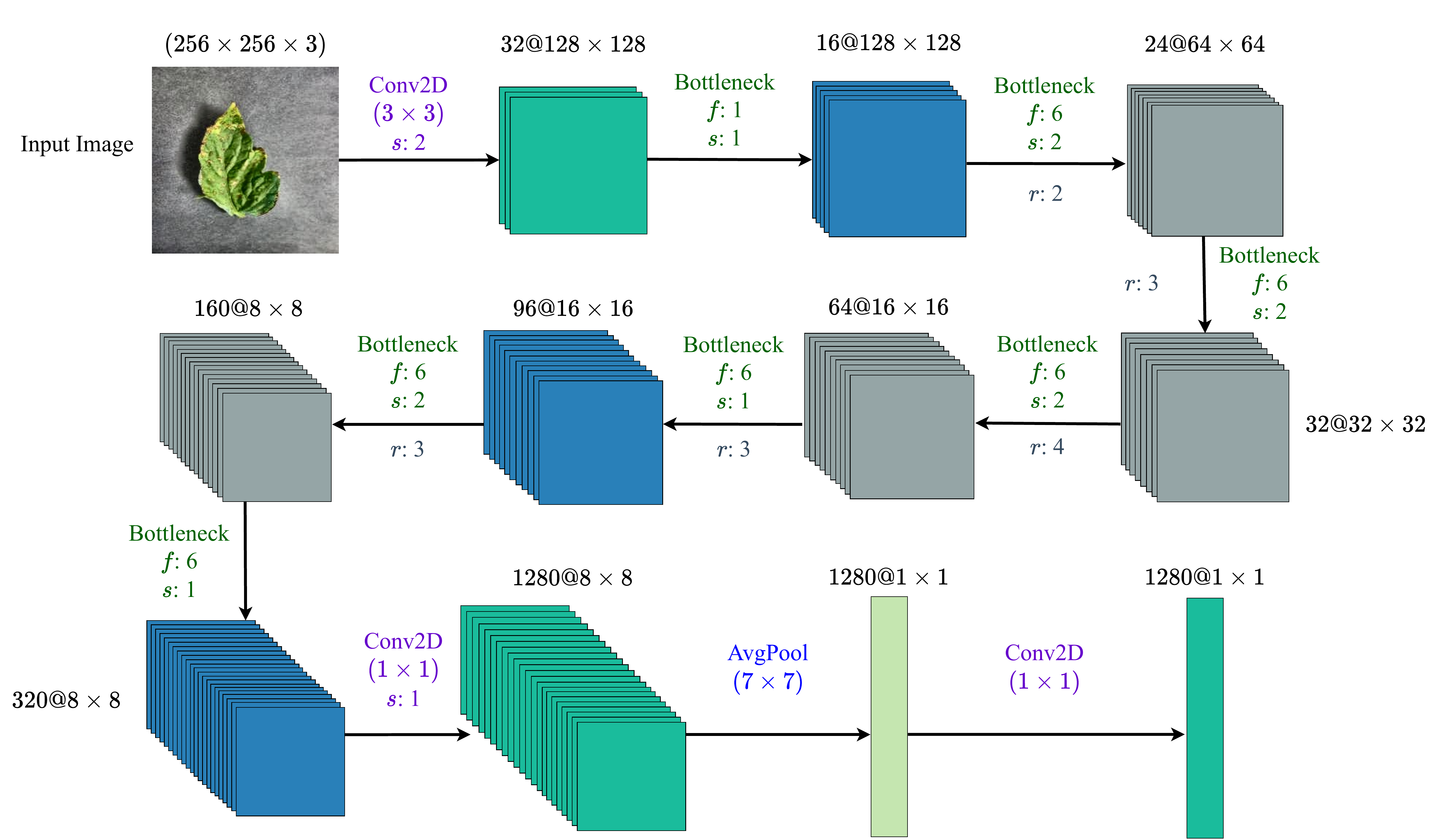}
    \caption{MobileNetV2 architecture adopted from \cite{sandler2018mobilenetv2} and modified for extracting features from $256\times256\times3$ tomato leaf images. Each box represents the feature maps (not to scale) after going through different layers. Here, $f$ denotes the expansion factor of each Bottleneck Layer. The first layer of each sequence has a stride value of $s$, and the remaining use stride 1. $r$ denotes the number of times a layer is repeated to produce the next feature map.}
    \label{fig:mobile}
\end{figure*}

Unlike traditional approaches in the existing literature on tomato leaf disease classification, we decided not to use data augmentations to increase the number of samples before training. Instead, these augmentations were performed randomly on different images during runtime in different splits, ensuring that the model sees different variations of the same image separately in different epochs. This reduces the possibility of overfitting, as it cannot see the same image in every epoch. On the other hand, this ensures that the different variations of the same image do not appear in both the training and the test set, thus eliminating the data leaking problem persistent in the existing literature.

\subsection{Transfer Learning-based Feature Extractor}

Earlier machine learning approaches assumed that the training and test data must be in the same feature space. However, recent advances in deep learning approaches have facilitated the use of an architecture trained to extract features on the training data of one domain to be used as a feature extractor for another domain. As the feature extractors in deep learning-based tasks became more and more generalized, this method of knowledge transfer, also known as Transfer Learning, has significantly improved the performance of learning, reducing a considerable amount of computational complexity. In this connection, the MobileNetV2 architecture has enabled real-time applications across multiple tasks and benchmarks using low computational resources. As shown in \figureautorefname~\ref{fig:mobile}, MobileNetV2 consists of a regular $3\times3$ convolution with 32 filters, followed by 17 Bottleneck Residual Blocks, a Pointwise convolution layer, a global average pooling layer, and a classification layer. The classification layer usually corresponds to the number of classes of the original dataset. For our system, the classification layer was replaced with a classifier network to classify tomato diseases.

\begin{figure*}[t]
    \centering
	\subfloat[Depthwise Separable Convolution with Stride-2 Block\label{fig:str2}]{
	\includegraphics[width=0.77\textwidth]{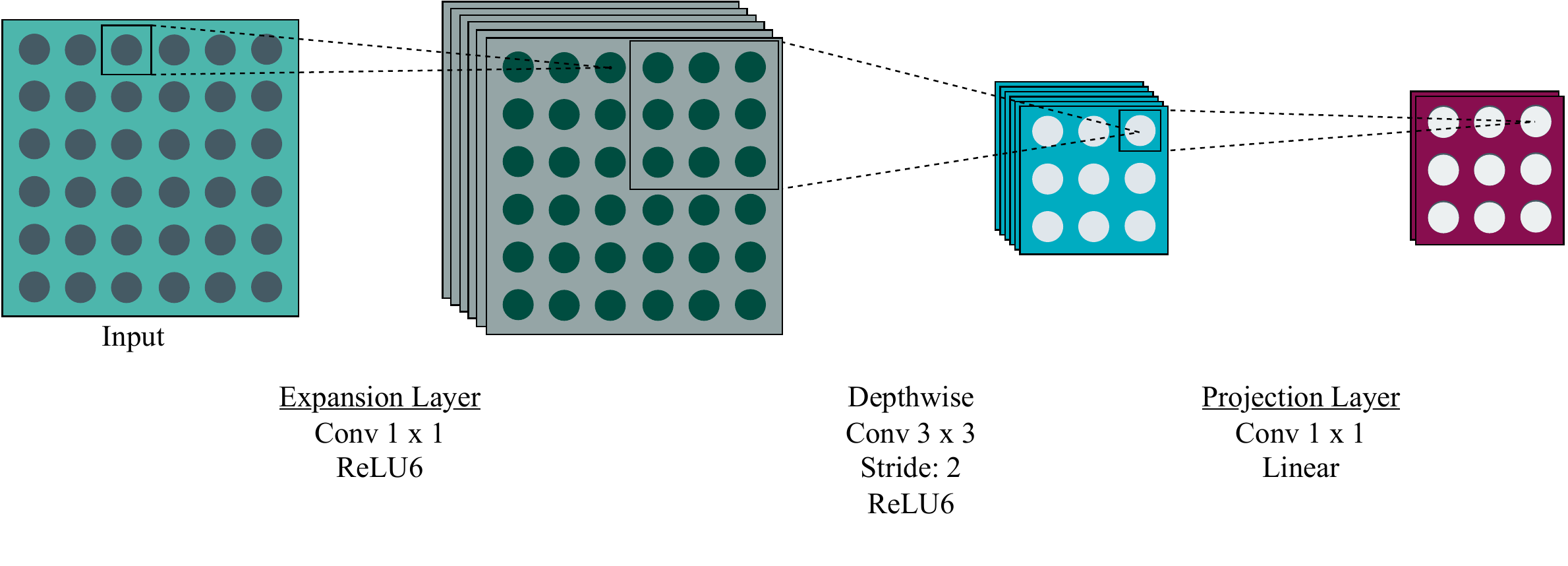}	
	}\\
    \subfloat[Inverted Residual Connection with Stride-1 Block\label{fig:str1}]{
	\includegraphics[width=0.77\textwidth]{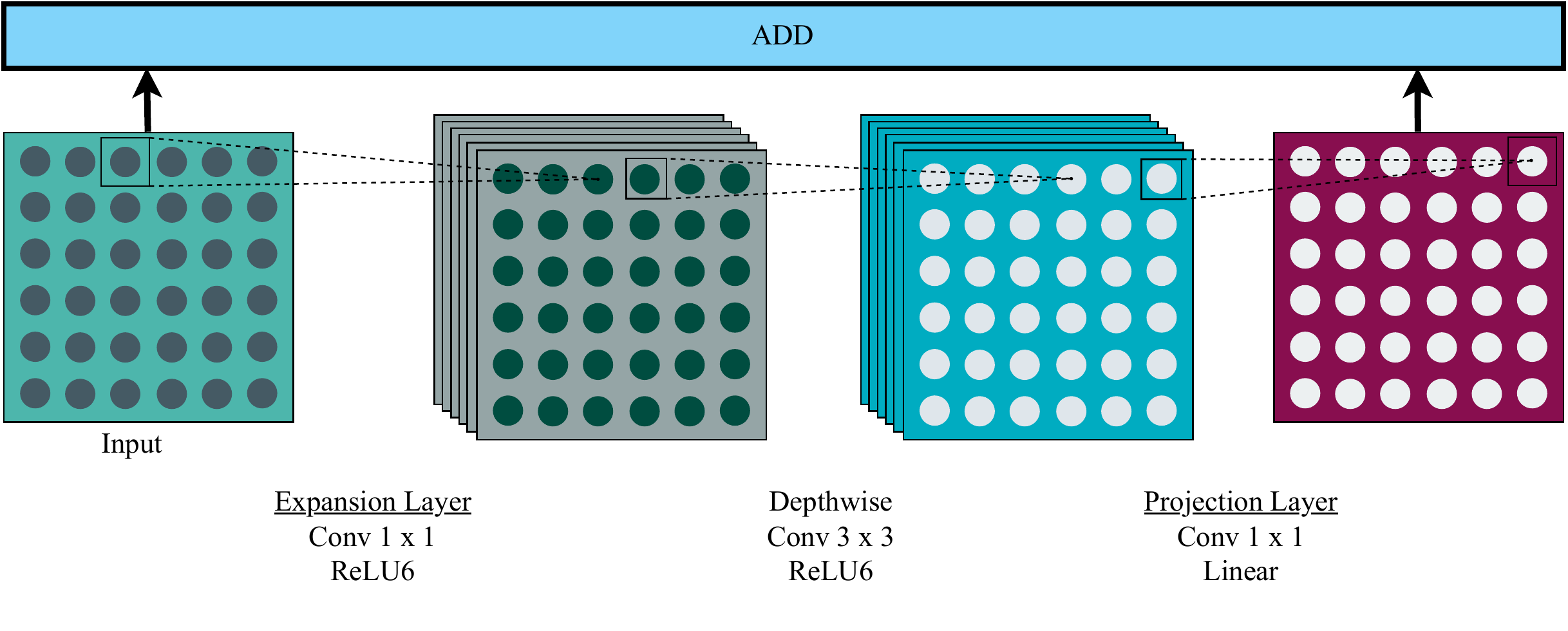}	
	}
    \caption{Bottleneck Residual Block. Here, each block represents the feature map output by different layers.}
\end{figure*}

At the heart of the MobileNetV2 architecture resides Bottleneck Residual Block containing three convolutional layers (\figureautorefname~\ref{fig:str2}). The Expansion Layer increases the number of channels in the input data by performing Pointwise convolution based on an expansion factor. The feature map output by this layer is then fed to a $3\times3$ Depthwise Convolution layer which works as a filter by applying convolution per channel. The Projection Layer takes these filtered values to generate salient features. Besides, this layer projects the higher dimensional data into a much lower number of dimensions, reducing the number of channels. The Depthwise Convolution layer combined with the Pointwise Convolution performs Depthwise Separable Convolution, reducing the computation by a factor of $O(k^2)$ compared to regular convolutions. Here, $k$ is the size of the Depthwise convolution kernel. Like most modern architectures, each of the three convolution layers is followed by batch normalization to stabilize the learning process. The activation function used by these layers is ReLU6. It bounds the activation within $[0, 6]$, making it more robust than the well-known ReLU function in fixed-point arithmetic. However, the Projection Layer does not contain any activation function due to the low dimensionality of the data produced by this layer. The non-linearity of the ReLU activation function can destroy valuable features. In addition, to reduce the effect of diminishing gradients, inverted residual connections are introduced through the network, which connects the bottleneck blocks with the same number of channels (\figureautorefname~\ref{fig:str1}).

In this work, to compare the performance of MobileNetV2, other transfer-learning architectures that are popular in leaf disease detection: DenseNet121, DenseNet201, EfficientNet-B0, MobileNet, NASNet-Mobile, ResNet50, ResNet\-152V2, and VGG19 were used.

\subsection{Classifier Network}
\begin{figure}[t]
    \centering
    \includegraphics[width=\columnwidth]{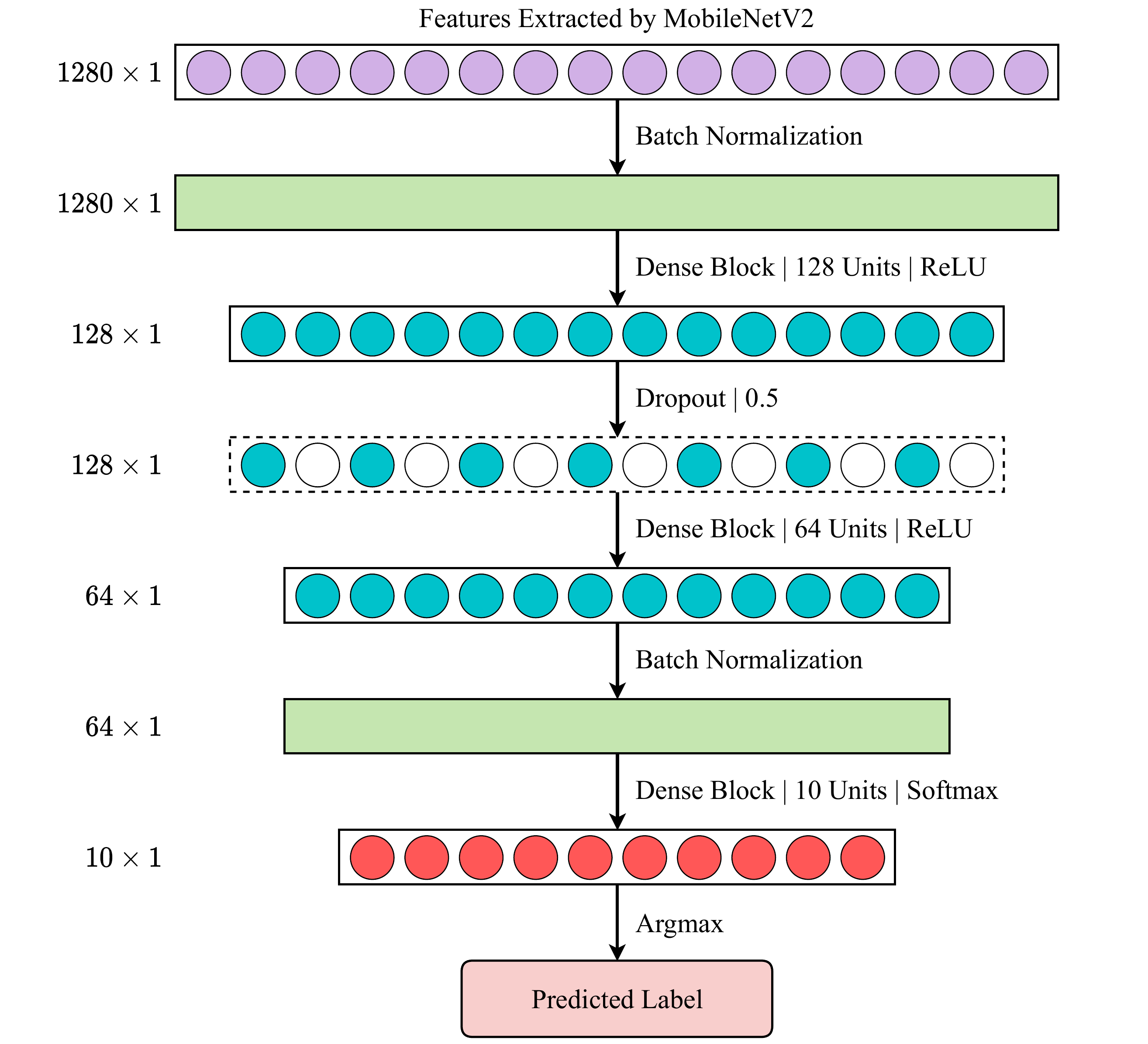}
    \caption{Classifier network}
    \label{fig:classifier}
\end{figure}

Instead of directly using the extracted features from pretrained models for final prediction, we employed a combination of dense, dropout, and batch normalization blocks to fine-tune the extracted traits further. As shown in \figureautorefname~\ref{fig:classifier}, dense blocks were added before the final output layer. The pretrained MobileNetV2 architecture we used was trained on a large and generalized dataset, making it perfect for feature extraction. The features extracted from the leaf images by MobileNetV2 architecture are then fed into the dense blocks trained from scratch to extract further the relevant features required to classify the diseases. Upon experimentation with different numbers of layers consisting of a varying number of nodes, two dense blocks with 128 and 64 nodes helped us achieve global minima in terms of loss.

A Batch Normalization \cite{iofee2015batch} block was added between the output of the MobileNetV2 and the first densely connected block and one between the second densely connected block and the output layer. A batch normalization block is used to standardize the inputs for the final layer for each mini-batch and stabilize the whole learning process, reducing the epochs needed to train the network.
Rectified Linear Unit (ReLU) \cite{glorot2011deep} was used as the activation function of the two densely connected blocks. This activation function makes the models easier to optimize and more generalizable. A dropout layer \cite{srivastava2014dropout} in-between these two dense blocks work as a regularizer, ensuring that the model does not overfit.
The final output layer of the classifier network is also a densely connected block with a Softmax activation function \cite{fellow2018softmax}.
The output layer of the classifier network contains ten nodes corresponding to each class label. The value of each node represents the probability of the input sample being in that class. Applying Argmax on this layer provides us with the predicted class label.

\subsection{Experimental Setup}
The proposed architecture was trained under a Python environment with TensorFlow, Keras, and other necessary libraries in Google Colab\footnote{\url{https://colab.research.google.com/}}. All experiments were conducted using an Intel Xeon CPU with a base clock speed of 2.3 GHz and an NVIDIA Tesla T4 GPU with a VRAM of 15 GB. The total usable memory of the machine was 13 GB.

From each class, the sample images were randomly split into $60\%$ for training, $20\%$ for validation, and $20\%$ for the test set. Following the mini-batch gradient descent technique \cite{khirirat2017minibatch}, the batch size was selected as 16. Since smaller batch sizes are often noisy, they help create a regularization effect and reduce the generalization error. They also help fit training data into memory.

While working with mini-batches, there is always a possibility of choosing batches that are not representative of the entire dataset, resulting in an inaccurate estimate of the gradient. Thus the training images were shuffled after each epoch throughout the experiments. This increases the probability that the model will converge and not be trapped with too many inferior batches.

For experimenting with mini-batches, it is crucial to shuffle the data after each epoch. The goal of shuffling data is to reduce variance and ensure that models generalize well and do not overfit.

The model was trained for at most 1000 epochs with early stopping. Early stopping helps reduce overfitting and improves the generalization of neural networks. Validation accuracy was selected as the scheme for evaluating the model so that early stopping can be triggered. In our proposed approach, a change in validation accuracy between epochs was considered significant if it was greater than $10^{\text{-}4}$. Otherwise, it was considered a patient epoch. The training was stopped early if there are ten consecutive patient epochs. Consequently, it was observed that our proposed architecture was able to converge after 70 epochs on average.

To ensure the rapid learning of salient features, we have used Adam optimizer \cite{diederik2015adam} for training our model. Compared to other optimizers, Adam can help multilayer deep learning networks converge faster for computer vision problems. Since this is a multiclass classification task, we have used categorical cross-entropy loss. The initial learning rate was set to $10^{\text{-}5}$. For every four consecutive patient epochs, the learning rate was decreased by a factor of 0.1 to help the model learn a set of globally optimal weights that leads to better optimization of the loss function.

The models can be initialized with different weights during the training phase, e.g., 0, random values, or pretrained weight values. In our work, we initialized the feature extractor part of the network with the respective pretrained weights from ImageNet Dataset \cite{olga2015imagenet} for the models and the classifier network with random weights. Model Checkpoints were used to save the model with the best validation accuracy so that they can be loaded later to continue the training from the saved state if required.

\subsection{Evaluation Metrics}
\subsubsection{Accuracy}
Accuracy is the ratio between the total number of predictions that were correct and the total number of predictions. To get a better estimation of the generalization capability of a model, the accuracy is calculated using the samples from the test set, which is unseen to the model during training.
\begin{align}
    \text{Accuracy} &= \frac{M}{N} \times 100 \%
\end{align}
Here, $N$ is the number of samples in the test set and $M$ is the number of samples for which the class labels were correctly predicted by the model.

\subsubsection{Parameter Count}
Parameters are the model's learnable weights, which are changed during the backward propagation phase based on the chosen optimization algorithm. The number of parameters can not only provide us with an idea regarding the training time of the model but also helps determine the model size and inference time.
\begin{align}
    \text{Parameter Count} &= \sum\limits_{i=1}^L p_i
\end{align}
Here, $p_i$ is the number of parameters in the $i$th layer and $L$ is the total number of layers in the model.

\subsubsection{Model Size}
Trained models can be stored as a Hierarchical Data Format version 5 (HDF5) file. The saved model contains the model's configuration, trained weights, and optimizer state. The model, along with its saved weights, can be loaded again to run inference. The size of the saved model is called the model size. Model size can be measured in MB (Megabyte) or GB (Gigabyte).

\subsubsection{FLOPs Count}
FLOPs Count is the theoretical maximum number of floating-point operations that a model requires to perform inference. Since the time taken for inference can vary from device to device, FLOPs Count is a better measurement to compare the relative inference time of deep learning models. It is usually measured in megaFLOPs (MFLOPs), gigaFLOPs (GFLOPs), or teraFLOPs (TFLOPs). The higher the value, the larger the number of computations required for a model to perform inference.
\subsubsection{Precision}
Precision is the ratio of the sum of the number of true positive predictions among all classes and the sum of the number of true positive predictions and false-positive predictions among all classes. In a multiclass problem, for each class, precision is used to evaluate the correctly classified samples of that class among all the samples that were classified as of that class. Precision is also called Positive Predictive Value (PPV).

Precision for each class $c$ can be calculated considering the one-vs-all strategy.

\begin{align}
    \text{Precision}_c &= \frac{TP_c}{TP_c+FP_c}
\end{align}
Here, $TP_c$ is the number of samples correctly classified as $c$, and $FP_c$ is the number of samples wrongly classified as $c$.

For imbalanced classes, macro-average precision is calculated where the precision for each class is calculated separately, and their average is taken. This ensures that the model gets equally penalized for each false positive instance of any class. 

For a set of classes $C$,

\begin{align}
    \text{Macro Average Precision} &= \frac{\sum\limits_{c\in C}\text{Precision}_c}{|C|}
\end{align}
Here, $\text{Precision}_c$ is the precision value for class $c$, and $|C|$ is the total number of classes.
\subsubsection{Recall}
Recall is the ratio of the sum of the number of true positive predictions among all classes and the sum of the number of true positive predictions and false-negative predictions among all classes. In a multiclass problem, recall is used to evaluate how many samples are correctly classified among all the samples that should have been classified as of that class. Recall is also called Sensitivity.

Recall for each class $c$ can be calculated considering the one-vs-all strategy.

\begin{align}
    \text{Recall}_c &= \frac{TP_c}{TP_c+FN_c}
\end{align}
Here, $TP_c$ is the number of samples correctly classified as $c$, and $FN_c$ is the number of samples of $c$ that are wrongly classified as other classes.

For imbalanced classes, macro average recall is calculated where the recall for each class is calculated separately and their average is taken. This ensures that the model gets equally penalized for each false-negative instance of any class. 

For a set of classes $C$,

\begin{align}
    \text{Macro Average Recall} &= \frac{\sum\limits_{c\in C}\text{Recall}_c}{|C|}
\end{align}
Here, $\text{Recall}_c$ is the Recall value for class $c$, and $N$ = Total number classes.

\subsubsection{F1-Score}
F1-Score is the weighted average of precision and recall that considers both the number of false positive predictions and false negative predictions. While working on an imbalanced dataset, having a high F1-Score is crucial to reduce the number of false positive and false negative predictions. 

F1-Score for each class $c$ can be calculated using the following formula:
\begin{align}
    \text{F1-Score}_c = \frac{2 \times P_c \times R_c}{P_c + R_c}
\end{align}
Here, $P_c$ is the precision value for class $c$, and $R_c$ is the recall value for class $c$.

For imbalanced classes, the macro average F1-score is calculated where the F1-score for each class is calculated separately and their average is taken. This ensures that each class gets equal priority in classification.

For a set of classes $C$,
\begin{align}
    \text{Macro Average F1-Score} &= \frac{\sum\limits_{c\in C}F_c}{|C|}
\end{align}
Here, $F_c$ is the F1-Score for class $c$, and $|C|$ is the total number of classes.

\begin{table*}[htb]
    \centering
    \caption{Comparison of the performance and characteristics among the baseline architectures on the original dataset}
    \label{tab:base}
    \begin{tabular}{L{0.16\textwidth}  C{0.1\textwidth}  C{0.12\textwidth}  C{0.08\textwidth}  C{0.1\textwidth}}
    \toprule
        Architecture & Accuracy (\%) & Parameters Count (Millions) & Model Size (MB) & FLOPs Count (MFLOPs)\\\midrule
        DenseNet121 & 97.96 & 7.1 & 27.58 & 14.1\\
        DenseNet201 & 99.36 & 18.35 & 71.11 & 36.69\\
        EfficientNet-B0 & 96.94 & 4.1 & 15.89 & 8.1\\
        MobileNet & 96.53 & 3.2 & 12.51 & 6.5\\
        MobileNetV2 & 97.27 & \textbf{2.28} & \textbf{8.98} & \textbf{4.54}\\
        NASNet-Mobile & 97.21 & 4.3 & 17.53 & 8.6\\
        ResNet50 & 98.70 & 23.62 & 98.29 & 51.11\\
        ResNet152V2 &  98.62 & 58.36 & 223.52 & 116.61\\
        VGG19 & \textbf{99.48} & 20.02 & 76.48 & 40.05\\\bottomrule
    \end{tabular}
\end{table*}

\subsubsection{AUC-ROC Score} 
Precision, recall, and the majority of the commonly utilized metrics have their individual set of restrictions. Precision is a measurement that determines how accurate a classification task is and it is only based on true positive and false positive predictions; a score of 1.0 for precision can be achieved with just one true positive prediction. 
On the other hand, recall is all about completeness and is solely based on true positive and false negative responses. As a result, predicting all the samples as positive will result in a recall of 1.0, but the precision will be very low. The Receiver Operating Characteristic (ROC) curve and the area under the ROC curve (AUC-ROC) are utilized as evaluation methods in this regard by combining the True Positive Rate (TPR) and False Positive Rate (FPR).

These methods allow models to be evaluated according to how well they separate classes from one another. The FPR and TPR for a series of predictions generated by the model at various thresholds are calculated to summarize the behavior of the model and can also be used to examine its ability to differentiate classes. Each probability threshold is represented by a point, linked to form a curve in the ROC graph. A model with no discriminating power is depicted by a diagonal line from FPR 0 and TPR 0 (coordinates: 0, 0) to FPR 1 and TPR 1. (co-ordinates: 1, 1).
%Below this line are models with a lower level of competency than none. 
A perfect model is represented as a point in the upper-left corner of the plot.

\section{Result and Discussion}
\label{sec:result}
For our experiments, we first investigated the performance of different baseline Deep CNN architectures to choose the best fit for our requirements. After that, an ablation study was conducted to justify how the different considerations in our proposed pipeline and modifications over the baseline contributed to improving our model's performance. Next, we inspected per-class precision, recall, and F1-score to evaluate how the proposed architecture addresses the class imbalance issue. Then, we compared the performance of our model with the existing state-of-the-artwork of tomato leaf disease classifications to establish its superiority. Finally, an error analysis was conducted to figure out where to invest the future improvement efforts.

\subsection{Performance of Different Baseline Architectures}
To choose the baseline model, several state-of-the-art Deep CNN architectures were implemented to perform tomato leaf disease classification. A comparison of their performance is shown in \tableautorefname~\ref{tab:base}.

The models were initialized with their pretrained weights on the ImageNet dataset and fine-tuned using the original tomato leaf samples from the PlantVillage dataset. The benefit of this initialization was that the models were already capable of learning complex patterns leading to faster convergence.
Since our goal was to pick the best-suited baseline for the proposed system, we only changed the final softmax layer with the number of classes of our dataset and trained without any enhancement or augmentations.

While choosing the appropriate architecture, we have considered the accuracy, number of trainable parameters, estimation of the number of floating-point operations (FLOPs), and the model's size. The VGG19 and DenseNet201 architectures achieved an accuracy higher than $99\%$ percent, and the performance of the ResNets also came close. These models are superior in terms of accuracy but have a significant disadvantage considering the other metrics.
For example, the VGG19 model has achieved 99.48\% accuracy, which is $2.2$\% higher than the accuracy of MobileNetV2 architecture. % 100*(99.48-97.27)/99.48
However, this improvement is costly in terms of memory and inference time. The model consumed $8.5$ times the storage space and $8.8$ times higher FLOPs count than MobileNetV2.  Similar can be said for DenseNet201 as well.
On the other hand, the relatively lighter models such as EfficientNet-B0, MobileNet, and NASNet-Mobile had lower accuracy than MobileNetV2 despite having higher values in terms of other metrics.

The MobileNetV2 architecture has the smallest model size and the lowest FLOPs count, making it ideal for real-time disease detection in devices with memory constraints. In addition to that, the fewer parameters of MobileNetV2 architecture result in faster training and inference. For these reasons, we chose MobileNetV2 as our base transfer learning architecture. We further aimed to improve the baseline performance by utilizing preprocessing techniques and an additional classifier network.

\subsection{Ablation Study}
\begin{table*}[hbt]
    \centering
    \caption{Ablation study of different components of the proposed pipeline}
    \label{tab:able}
    \begin{tabular}{C{1cm} C{2.4cm} C{2.8cm} C{1.2cm}}
        \toprule
        CLAHE & Augmentation & Classifier Network & Accuracy\\\midrule
        \xmark & \xmark & \xmark & 97.27\\
        \xmark & \xmark & \checkmark & 98.29\\
        \xmark & \checkmark & \xmark & 98.46\\
        \xmark & \checkmark & \checkmark & 99.03\\
        \checkmark & \xmark & \xmark & 97.71\\
        \checkmark & \xmark & \checkmark & 98.60\\
        \checkmark & \checkmark & \xmark & 98.84\\
        \checkmark & \checkmark & \checkmark & 99.30\\
        \bottomrule
    \end{tabular}
\end{table*}

An ablation study was conducted to understand the contribution of different components of the proposed pipeline to the overall performance.
We considered several combinations of the design choices like the preprocessing steps, such as CLAHE, data augmentation, and the introduction of a classifier network to analyze their effects. A summary of the result in different settings can be found in \tableautorefname~\ref{tab:able}.

A positive impact can be seen on the results when the images are preprocessed using CLAHE. This can be attributed to CLAHE for enhancing the leaf images' disease spots, making them more prominent and easier to identify for the models. For example, the baseline performance of $97.27\%$ was improved to $97.71\%$ after we introduced CLAHE.
We noticed a further improvement in the results when data augmentation was introduced. The runtime augmentations allow the model to learn from different representations of the images in every epoch, allowing the model to focus on the features highlighted by CLAHE.

Experiments were performed to find out how data augmentation in different splits affects the overall performance. 
We found that performing data augmentation on all three splits resulted in the best accuracy. As a result of the augmentation, the model learns to recognize different variations of the original image during the training phase. However, without augmenting the test set, no such variations are to be found. This violates the key assumption in dataset splitting for general classification tasks, that the distribution of images found in the training and validation set should be similar to the distribution in the test set.

One key factor here is that, as all our samples are being augmented with random probability during run-time, the model never sees the same version of an image twice. Augmentation in training and validation splits ensures that the model hardly gets any chance to overfit and learns generic feature representations. In addition, augmentations performed on the test set ensure that those samples represent real-life scenarios, making the classification task even more challenging. However, this begets a problem. Since the augmentation is performed randomly, the model sees different images in each test run. As a result, the accuracy for each run might not be the same; instead, it gives us a value within a range.
To resolve this issue, we tested the trained model 100 times whenever we used augmentation and reported the average accuracy.
The benefits of doing this are two-fold. First, as the test set is randomly augmented, the average accuracy is a better descriptor of the model's performance, preventing any chance of getting lucky. Moreover, these trials are testing the model with a variety of samples, more than what could be done using a static test set. So a model being able to do well in this setup will be robust and can be expected to achieve similar accuracy in real-life scenarios. It is worth mentioning that the maximum accuracy achieved by our best model was $99.53\%$.

% Effect of Classifier network
Our hypothesis of introducing the classifier network was that the model would be able to consider further combinations of the extracted features from the MobileNetV2 network, leading to improved overall performance.
Since this network was trained from scratch on the provided information from the feature extractor network, it extracted even more meaningful features for leaf disease classification. Thus we found an improvement in the overall accuracy every time the classifier network was introduced in different setups.

Initially, the performance of the baseline MobileNetV2 model was only 97.27\%. The combination of the preprocessing techniques increased it up to 98.84\% showing how these choices improve the generalizing capability of the model. Finally, the model’s competence was further enhanced with the classifier network leading to a mean accuracy of 99.30\% (Standard Deviation: $0.00095$) over 100 runs.

\subsection{Addressing the Class Imbalance}
As mentioned earlier, there exists a class imbalance in the PlantVillage dataset. Thus drawing a conclusion on a model's performance solely based on the accuracy metric might be unwise as the accuracy might still be in the $90^{th}$ percentile even if the model is incapable of classifying half of the samples of the least populated classes. To tackle this issue, macro-averaged precision, recall, and F1-score values were taken under consideration, which gives equal importance to all the classes regardless of the number of samples. Our proposed model achieves $99.18$ precision, $99.07$ recall, and $99.12$ F1-score. The high values of precision and recall signify that our model does a great job identifying the True Positives. At the same time, it penalizes the accidental False Positive and False Negative cases. Taking a harmonic mean of these two metrics, the $99.12$ value of the F1-score proves the robustness of the proposed architecture even in imbalanced datasets.

\begin{table*}[htb]
    \centering
    \caption{Per class precision, recall, F1-Score for the test set}
    \label{tab:perclass}
    \begin{tabular}{L{.23\textwidth} C{.1\textwidth}  C{.1\textwidth} C{.1\textwidth} C{.1\textwidth}}
    \toprule
        Class Label & Sample Count & Precision & Recall & F1-Score\\
    \midrule
        Bacterial Spot & 425 & 0.9976 & 0.9953 & 0.9965\\
        Early Blight & 200 & 0.9745 & 0.9550 & 0.9646\\
        Late Blight & 381 & 0.9794 & 0.9974 & 0.9883\\
        Leaf Mold & 190 & 1.000 & 0.9895 & 0.9947\\
        Septoria Leaf Spot & 354 & 0.9915 & 0.9915 & 0.9915\\
        Two-spotted Spider Mite & 335 & 0.9852 & 0.9940 & 0.9896\\
        Target Spot & 281 & 0.9928 & 0.9858 & 0.9893\\
        Yellow Leaf Curl Virus & 1071 & 1.0000 & 0.9981 & 0.9991\\
        Tomato Mosaic Virus & 74 & 1.0000 & 1.0000 & 1.0000\\
        Healthy & 318 & 0.9969 & 1.0000 & 0.9984\\\bottomrule
    \end{tabular}
\end{table*}

Furthermore, 
\tableautorefname~\ref{tab:perclass} shows the precision, recall, and F1-score for each class. From the table, it is evident that our data augmentation technique solved the class imbalance problem as these values are high even for the classes with a low number of samples.

\subsection{Class Separability}

The AUC-ROC curve can assess the performance of a predictive model by describing the trade-off between the True Positive Rate and False Positive Rate by employing multiple probability thresholds. A perfect classifier will have a ROC where the graph reaches 100\% true positives and zero false positives. We generally measure the number of positive classifications with an increment in the rate of false positives. 

\begin{figure}[htb]
    \centering
    \includegraphics[width=\columnwidth]{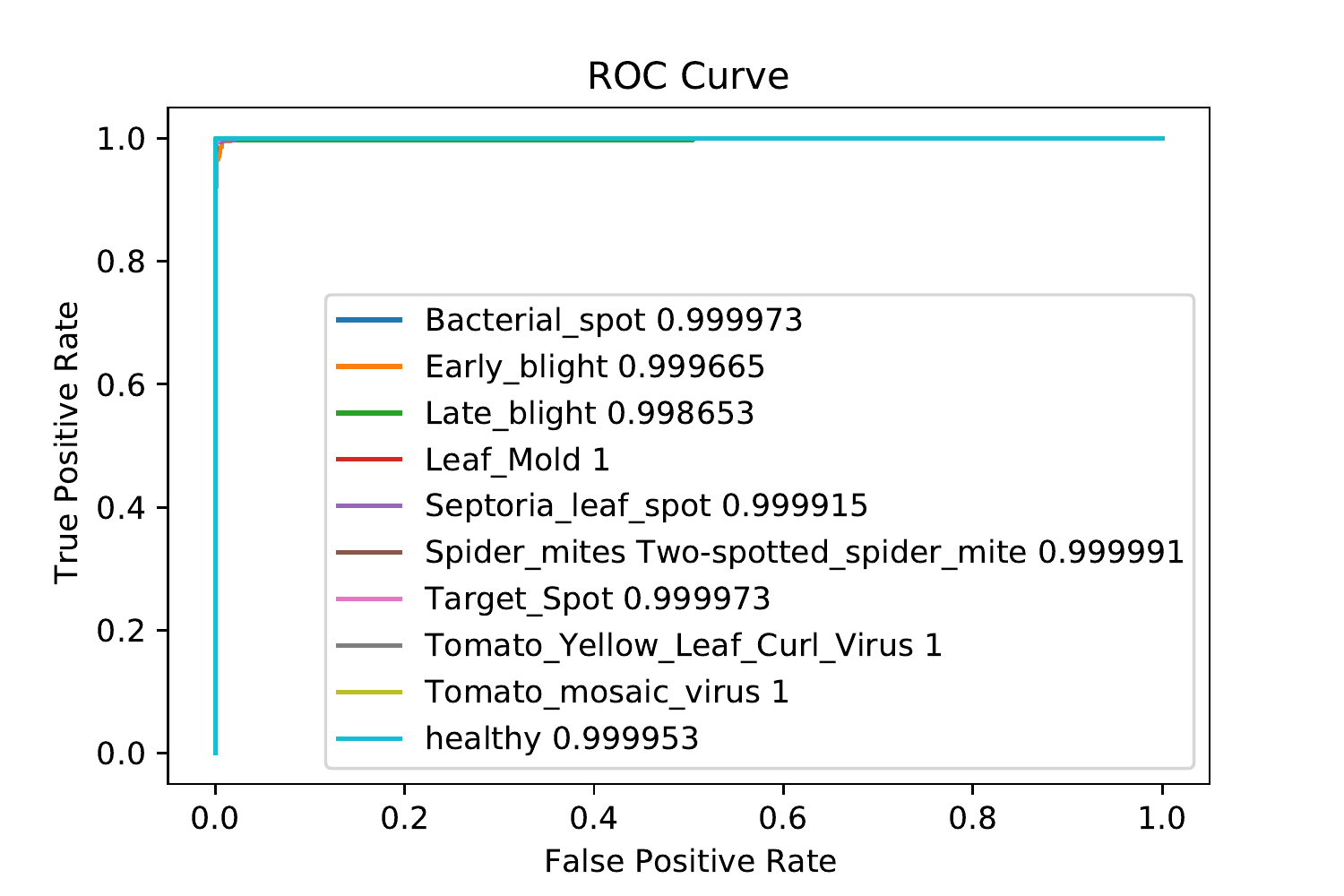}
    \caption{ROC curve for the tomato leaf diseases}
    \label{fig:classwiseROC}
\end{figure}

As shown in \figureautorefname~\ref{fig:classwiseROC}, the ROC curves overall each other at the top-left corner. That means, our proposed architecture achieves a commendable performance for all 10 classes. Among the classes, our model achieved an AUC score of 1 for Leaf Mold, Yellow Leaf Curl Virus, and Mosaic Virus. Scores for the other classes are also fairly high, indicating a satisfactory class separability.
% illustrates the ROC curves for each disease class individually. Leaf Mold, Yellow Leaf Curl Virus, and Mosaic Virus scored a perfect 1 for class separability, whereas Early Blight scored 0.999665.

\subsection{Comparison with State-of-the-art Methods}
\tableautorefname~\ref{tab:sota} presents a comparison of our proposed architecture with the state-of-the-art models of tomato leaf disease classification. Our model achieves a commendable accuracy of $99.30\%$ while keeping the model size and the number of operations low. Comparing it to the state-of-the-art models, we can notice that only \cite{maeda2020comparison} achieved a mere $0.09\%$ increase in accuracy, having 2.4 times the model size and $59.27\%$ increase in FLOPs count. Our model's smaller size and low computational cost without sacrificing performance make it suitable for low-end devices.

\begin{table*}[!hbt]
    \centering
    \caption{Performance comparison with the State-of-the-Art models for tomato leaf disease classification}
    \label{tab:sota}
    \begin{tabular}{L{.22\textwidth}  C{0.05\textwidth}  C{0.06\textwidth}  C{0.07\textwidth}  C{0.08\textwidth}  C{0.1\textwidth}}
        \toprule
        Reference & Class Count & Image Count & Accuracy (\%) & Model Size (MB) & FLOPs Count (MFLOPS)\\
        \midrule
        Durmu\c{s} \etal \cite{durmucs2017disease} & 10 & N/A & $94.30$ & $2.94$ & $1.44$\\
        Brahimi \etal \cite{brahimi2017deep} & 9 & 14828 & $99.18$ & $23.06$ & $11.95$\\
        Tm \etal \cite{tm2018tomato} & 10 & 18160 & $ 94.85$ & $156.78$ & $82.18$\\
        Rangarajan \etal \cite{rangarajan2018tomato} & 7 & 13262 & $97.49$ & $350.25$ & $183.53$\\
        Zhang \etal \cite{zhang2018can} & 9 & 5550 & $97.28$ & $98.29$ & $51.1$\\
        Bir \etal \cite{bir2020transfer} & 10 & 15000 & $98.60$ & $15.59$ & $8.11$\\
        Maeda-Gutierrez \etal \cite{maeda2020comparison} & 10 & 18160 & $99.39$ & $23.06$ & $11.95$\\
        Wu \etal \cite{wu2020dcgan} & 5 & 5300 & $94.33$ & $23.06$ & $11.95$\\
        Abbas \etal \cite{abbas2021tomato} & 10 & 16012 & $97.11$ & $27.58$ & $28.09$\\
        \midrule
        Proposed Method & 10 & 18160 & $99.30$ & $9.60$ & $4.87$\\
        \bottomrule
    \end{tabular}
\end{table*}

\begin{figure*}[!htb]
    \centering
    \subfloat[Accuracy vs. Model Size\label{fig:model}]{
    \includegraphics[width=\textwidth]{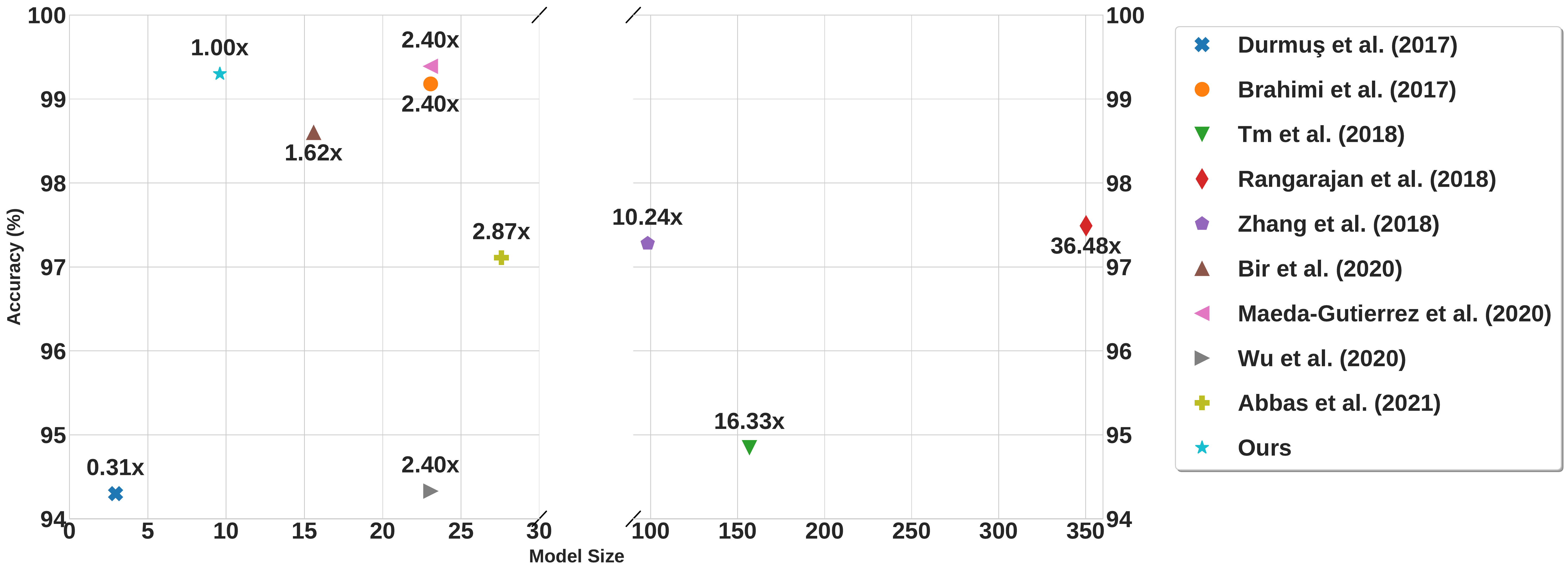}
    }\\
    \subfloat[Accuracy vs. FLOPs Count\label{fig:flops}]{
    \includegraphics[width=\textwidth]{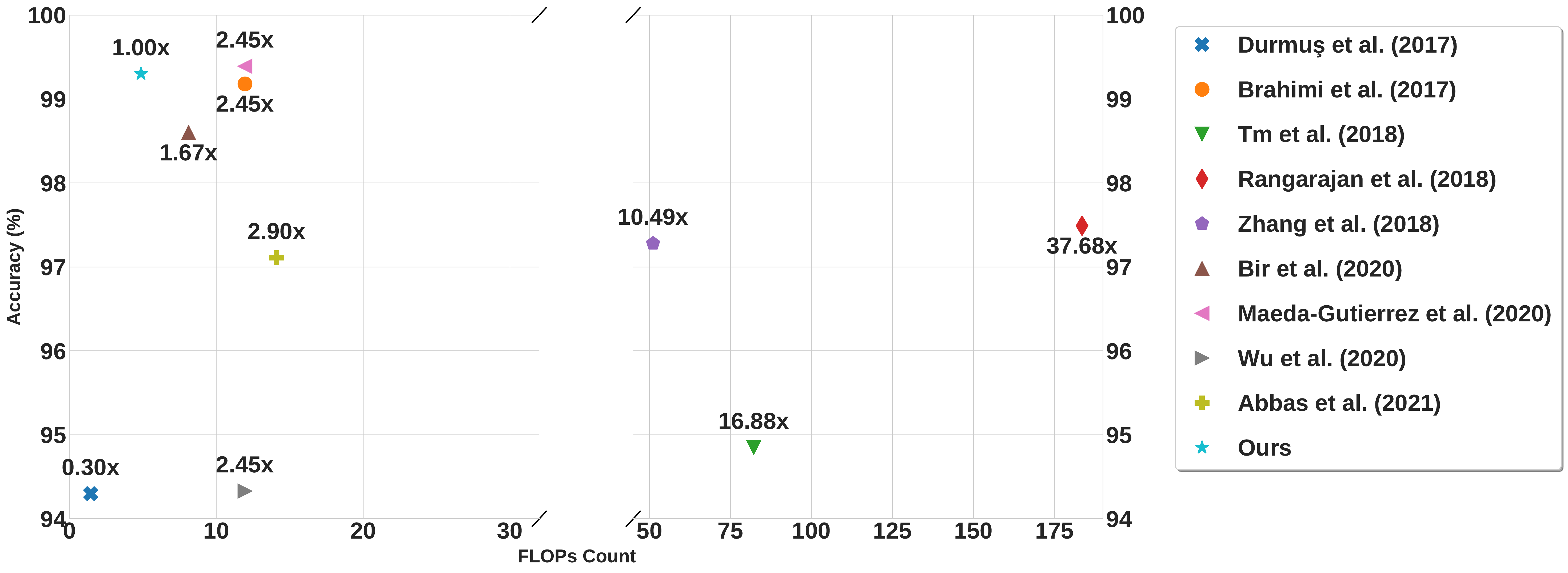}
    }
    \caption{Performance comparison with state-of-the-art tomato leaf disease classification architectures based on model size and FLOPs count} 
\end{figure*}

Some of the works mentioned in the table did not utilize all the samples from the subset of the PlantVillage dataset, which leaves a possibility of accidentally missing some critical samples \cite{zhang2018can, bir2020transfer, wu2020dcgan, abbas2021tomato}. Additionally, some of the models did not consider all the classes, which might lead to the misclassification of unseen samples. For example, \cite{brahimi2017deep} achieved an accuracy of $99.18\%$, but the experiment did not include any healthy samples of tomato leaves. This results in labeling a healthy leaf sample to any of the disease classes.

% Discussion on Model Size vs Acc vs Flops
Further analysis shows that the space requirement of our proposed architecture is only 9.6MB. In contrast, different works in the existing literature required at least twice of this storage space, if not more, to produce similar accuracy (\figureautorefname~\ref{fig:model}). Although \cite{durmucs2017disease} has a smaller model size than that of ours, the accuracy is far less.
\figureautorefname~\ref{fig:flops} shows that our model significantly reduced the FLOPs requirement without compromising the accuracy. Hence it removes the requirement for high-performance hardware along with reducing the inference time of the model. 
It can be observed that despite using deeper models, some works could not achieve comparable performance to the state-of-the-art. This further justifies the usefulness of the different components of our proposed architecture.

\subsection{Qualitative Analysis}

\subsubsection{Focusing on the Diseased Portion}
\begin{figure}[htb]
    \centering
    \subfloat[GradCam output for Bacterial Spot\label{fig:quantBact}]{
    \includegraphics[width=0.45\linewidth]{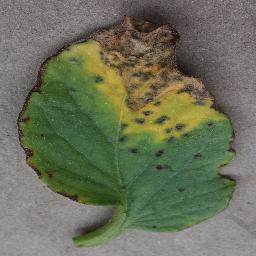} \hfill \includegraphics[width=0.45\linewidth]{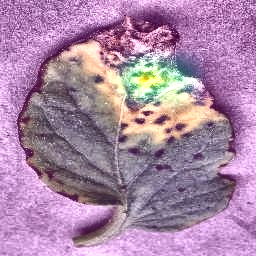}
    }\\
    
    \subfloat[GradCam output for Late Blight\label{fig:quantLate}]{
    \includegraphics[width=0.45\linewidth]{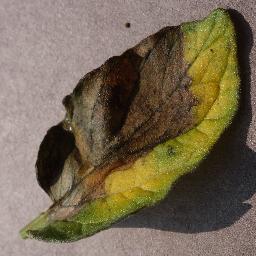} \hfill \includegraphics[width=0.45\linewidth]{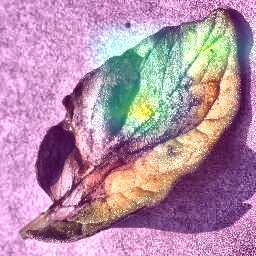}
    }\\
    \subfloat[GradCam output for Healthy Leaf\label{fig:quantHeal}]{
    \includegraphics[width=0.45\linewidth]{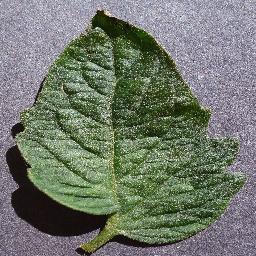} \hfill \includegraphics[width=0.45\linewidth]{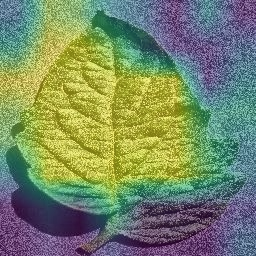}
    }
    \caption{Qualitative results showing GradCam output for correctly classified samples}
    \label{fig:gradcamCorr}
\end{figure}

To get an intuitive understanding of whether our model is learning to correctly predict considering the relevant features or not, we examine the Gradient-weighted Class Activation Mapping (GradCAM) output for the correctly classified samples \cite{selvaraju2017gradcam} (\figureautorefname~\ref{fig:gradcamCorr}). This visualization can show us how our model classifies a diseased or healthy sample by highlighting the region in which our model focuses while making the class label decision.

As shown in \figureautorefname~\ref{fig:gradcamCorr}\subref{fig:quantBact} and \ref{fig:gradcamCorr}\subref{fig:quantLate}, our model focused on the diseased portion of the leaf image to provide the class decision. Conversely, in the case of a healthy leaf (\figureautorefname~\ref{fig:gradcamCorr}\subref{fig:quantHeal}), the model focuses on the entire leaf image finding no diseased portion and then classifies the image as healthy. This goes to show the capability of our model in understanding what to look for in a leaf image to make the correct class decision.

\subsubsection{Error Analysis}
\label{sub:error}

\begin{figure}[htb]
    \centering
    \includegraphics[width=0.95\columnwidth, trim={0 0 1cm 0},clip]{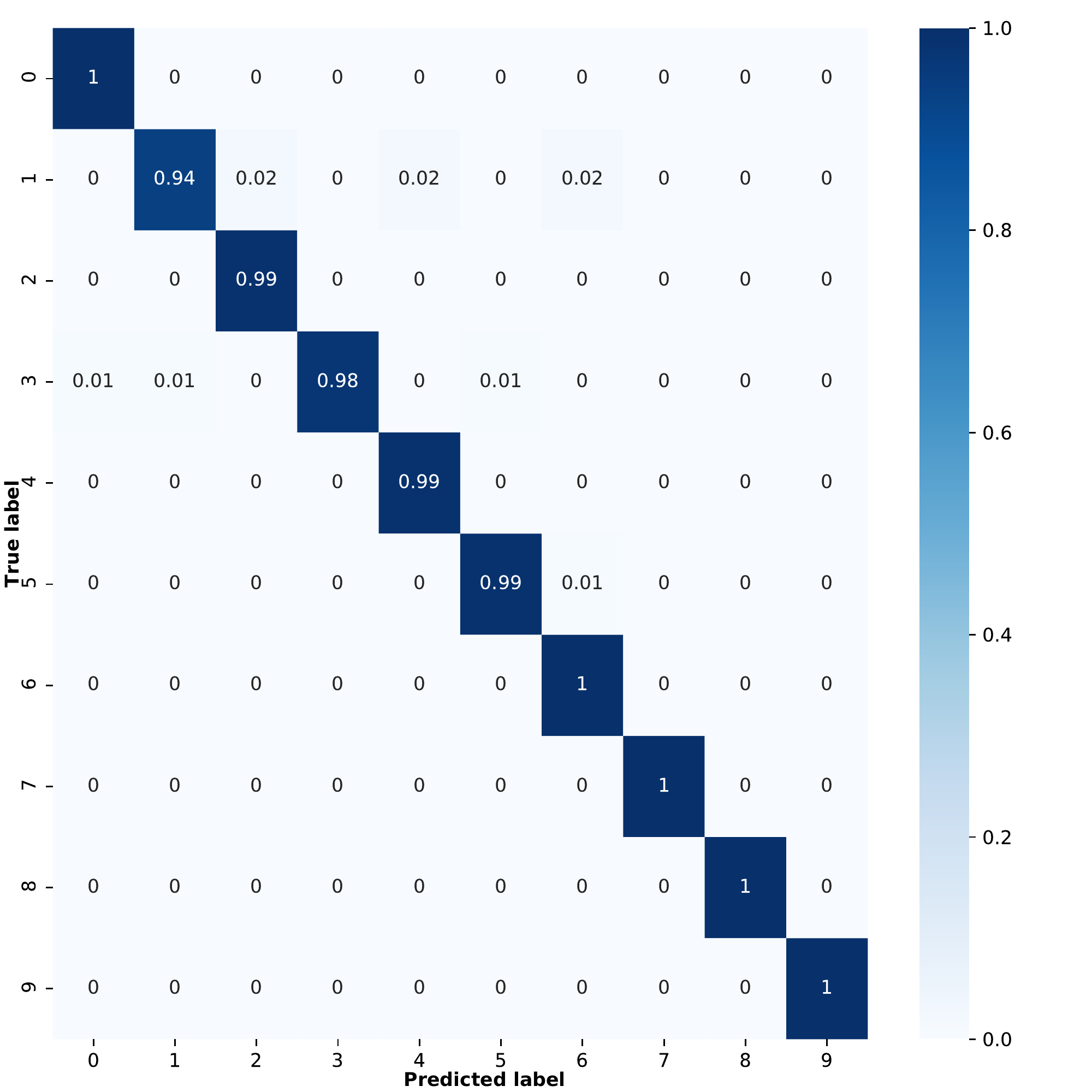}
    \caption{Confusion matrix. Here, 0 = Bacterial spot, 1 = Early blight, 2 = Late Blight, 3 = Leaf Mold,  4 = Septoria Leaf Spot, 5 = Two-spotted Spider Mite, 6 = Target Spot, 7 = Yellow Leaf Curl Virus, 8 = Tomato Mosaic Virus, 9 = Healthy}
    \label{fig:confmat}
\end{figure}

\begin{figure}[htb]
    \centering
	\subfloat[Misclassified Early Blight Sample and its GradCAM output\label{fig:misearly}]{
	\includegraphics[width=.45\columnwidth]{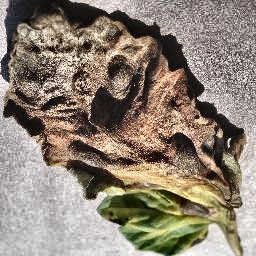} \hfill \includegraphics[width=0.45\columnwidth]{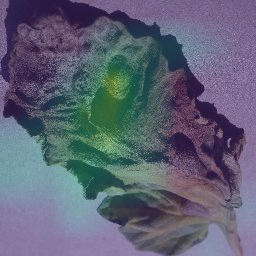}
	}\\
    \subfloat[Similar Late Blight Samples\label{fig:mislate}]{
	\includegraphics[width=.45\columnwidth]{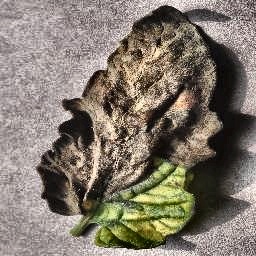}\hfill
	\includegraphics[width=.45\columnwidth]{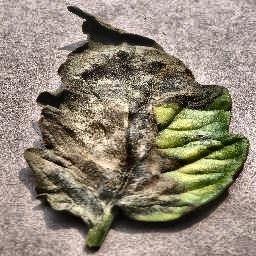}
	}
    \caption{Misclassified sample with visually similar samples of the predicted class}
    \label{fig:error}
\end{figure}

According to the confusion matrix of our best performing model (\figureautorefname~\ref{fig:confmat}), for half of the classes, our model was able to predict all the unseen test samples correctly. For the rest, the accuracy is comparable to other state-of-the-art methods. However, the most misclassified samples were from the `Early Blight' class. A few of the misclassified samples from this class were predicted as `Late Blight'. Upon reviewing the misclassified samples, we identified visually similar leaves from both classes. 
For example, in the original dataset, the class label for \figureautorefname~\ref{fig:error}\subref{fig:misearly} is `Early Blight', which was misclassified to `Late Blight' class during inference. The GradCAM output for the sample shows that our model correctly focuses on the diseased region of the leaf image (\figureautorefname~\ref{fig:error}\subref{fig:misearly}).

However, in the training set of the Late Blight class, there are several images (e.g. \figureautorefname~\ref{fig:error}\subref{fig:mislate}) that are similar to \figureautorefname~\ref{fig:error}\subref{fig:misearly}. Since during training, the model learns to classify these images as of the class `Late Blight', it is expected that similar images from the test set will also be classified in the same class.

To conclude, after analyzing the misclassified samples, we found some inter-class similarities in the infected regions among some of the diseases. A few of the leaves were severely damaged by the virus, which eventually restricted the model from extracting meaningful features leading to misclassification.

\section{Conclusions}
\label{sec:conclusions}
%Human civilization needs to increase food production, keeping pace with population growth where yield loss due to different leaf diseases is one of the biggest obstacles. Fast and accurate recognition of diseases and proper diagnosis can go a long way to solve this challenge. 
Fast and accurate recognition of leaf diseases can go a long way to meeting the ever-increasing demand in food production. In this regard, we have proposed a lightweight deep neural network by combining a fine-tuned pretrained model and a classifier network. The utilization of the adaptive contrast enhancement technique has eliminated the illumination problem persistent in the dataset. Runtime data augmentation techniques have been applied to address the class imbalance issue while avoiding data leakage. All these components of the pipeline enabled the model to focus on the disease spots and extract high-level features leading to an accuracy of $99.30\%$. We achieved this performance with a significantly smaller model size and FLOPs count compared to the state-of-the-art models. This makes the proposed pipeline a suitable choice for building applications for low-end devices.
% With a requirement of $9.60$ MB memory space and $4.87$M floating-point operations, the proposed model achieves $99.30\%$ accuracy, making it a suitable choice for 
%With a high accuracy and F1-score, this relatively smaller and faster model can be used in real-life applications in devices with less memory and low computational power to precisely predict diseases from tomato leaves.

However, one of the limitations of the PlantVillage dataset is that the samples are taken in laboratory conditions. Further experiments can be performed with tomato leaf images with varying backgrounds taken from the field. Such images might contain occlusion and background clutter. Advanced segmentation techniques can be taken into account to locate the infected regions before classification. Moreover, only a single disease can be found in each of the samples used in our experiment. Identifying multiple diseases within a single leaf will be another challenging task to solve. The classification goal can also include detection of the severity of infection on leaves, which intelligent systems can utilize to decide the amount of pesticide to be used. Finally, this work can be extended to classify diseases from a broader range of crops.

\section*{Acknowledgment}
The authors are thankful to A.B.M. Ashikur Rahman, Mohammad Ishrak Abedin, and Shahriar Ivan, Department of Computer Science and Engineering, Islamic University of Technology, and Mst. Nura Jahan, Department of Entomology, Bangabandhu Sheikh Mujibur Rahman Agricultural University for their valuable time and support in making this study possible.

\section*{Conflict of Interest}
The authors declare that there is no conflict of interest.

\bibliographystyle{IEEEtran}
\bibliography{ref}

\begin{IEEEbiography}[{\includegraphics[width=1in,height=1.25in,clip,keepaspectratio]{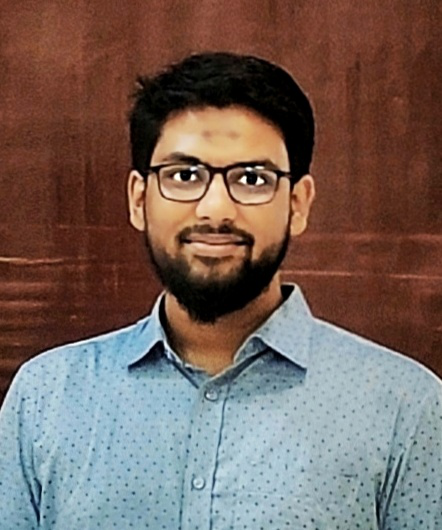}}]{Sabbir Ahmed} was born in Dhaka, Bangladesh, in 1996. He graduated from Islamic University of Technology (IUT), Gazipur, Bangladesh, in 2017 with a B.Sc. Engg. degree (IUT Gold Medalist) in Computer Science (CS) and is pursuing his M.Sc. degree in CS from the same institution.

Since 2018, he has been working as a Lecturer in the Department of Computer Science and Engineering, IUT. His research interests include pattern recognition, deep learning in computer vision, and intelligent agriculture.
\end{IEEEbiography}

\begin{IEEEbiography}[{\includegraphics[width=1in,height=1.25in,clip,keepaspectratio]{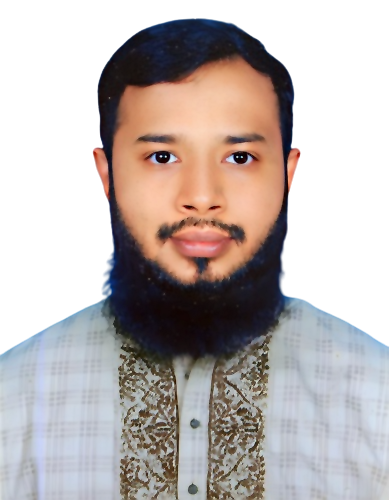}}]{Md. Bakhtiar Hasan} completed his B.Sc. Engg. and M.Sc. Engg. degree in Computer Science and Engineering (CSE) from Islamic University of Technology (IUT) in 2018 and 2022, respectively.

Since 2019, he has been working as a Lecturer in the Department of Computer Science and Engineering, IUT. His research interest includes the use of deep learning and computer vision techniques in human biometrics and smart agriculture.

Mr. Hasan received Huawei Seeds for the Future scholarship in 2018. He was awarded IUT Gold Medal in recognition of his outstanding performance in the pursuit of the B.Sc. Engg. in CSE degree in 2018.
\end{IEEEbiography}

\begin{IEEEbiography}[{\includegraphics[width=1in,height=1.25in,clip,keepaspectratio]{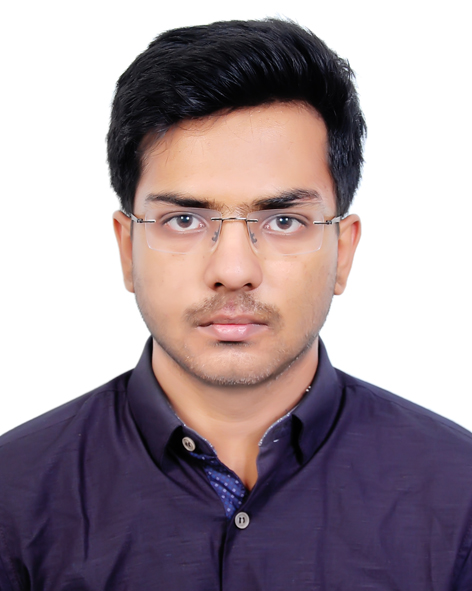}}]{Tasnim Ahmed} was born in Kushtia, Bangladesh in 1997. He received his B.Sc degree in Computer Science and Engineering from the Islamic University of Technology, Gazipur, Bangladesh and currently he is pursuing the M.Sc degree.

Since 2020, he has been working as a full-time Lecturer with the Computer Science and Engineering Department, Islamic University of Technology, Gazipur, Bangladesh. His research interests include computer vision, natural language processing, bioinformatics, and software engineering.
\end{IEEEbiography}

\begin{IEEEbiography}[{\includegraphics[width=1in,height=1.25in,clip,keepaspectratio]{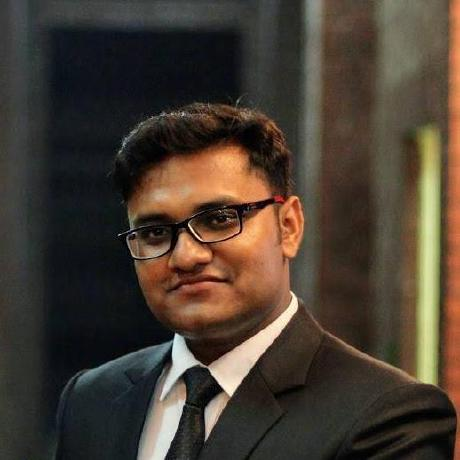}}]{Redwan Karim Sony} received his B.Sc. in Computer Science and Engineering from Islamic University of Technology (IUT) in 2016. From 2017 to 2021, he served as a Lecturer in the Department of Computer Science and Engineering, IUT. He is now on study leave and currently pursuing his Ph.D. at Michigan State University in the USA. His research includes facial biometrics, model visualization, and the explainability of deep learning models.
\end{IEEEbiography}

\begin{IEEEbiography}[{\includegraphics[width=1in,height=1.25in,clip,keepaspectratio]{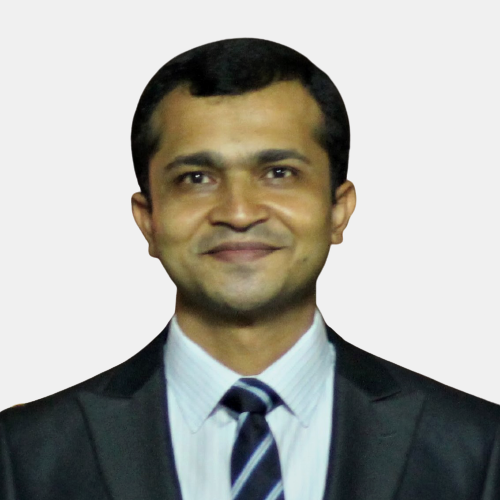}}]{Md. Hasanul Kabir} (M’17) received the B.Sc. degree in computer science and information technology from the Islamic University of Technology, Bangladesh, and the Ph.D. degree in computer engineering from Kyung Hee University, South Korea.

He is currently a Professor in the Department of Computer Science and Engineering, Islamic University of Technology. His research interests include feature extraction, visual question answering, and sign language interpretation by combining image processing, machine learning, and computer vision.
\end{IEEEbiography}

\EOD

\end{document}